\documentclass{article}
\usepackage[normalem]{ulem}
\useunder{\uline}{\ul}{}
\usepackage{PRIMEarxiv}  
\usepackage[utf8]{inputenc}
\usepackage[table]{xcolor}
\usepackage{enumitem}
\usepackage{array}
\usepackage{url}        
\usepackage{float}

\usepackage{booktabs}   
\usepackage{amsfonts}   
\usepackage{nicefrac}   
\usepackage{longtable}
\usepackage{microtype}  
\usepackage{lipsum}     
\usepackage{graphicx}   
\graphicspath{{media/}} 
\usepackage{caption}    
\usepackage{amsmath}

\usepackage{array}       
\usepackage{booktabs}    
\usepackage{multicol}    
\usepackage{hyperref}  

\title{\textnormal{Samba-ASR: State-Of-The-Art Speech Recognition Leveraging Structured State-Space Models}}

\author{\textbf{Syed Abdul Gaffar Shakhadri}\\
\small{Lead AI Developer}\\
\small{SandLogic Technologies Pvt Ltd.}\\
\small{\texttt{syed.abdul@sandlogic.com}}
\and
\textbf{Kruthika KR}\\
\small{AI Researcher}\\
\small{SandLogic Technologies Pvt Ltd}\\
\small{\texttt{kruthika.kr@sandlogic.com}}
\and
\textbf{Kartik Basavaraj Angadi}\\
\small{AI Developer}\\
\small{SandLogic Technologies Pvt Ltd}\\
\small{\texttt{kartik.angadi@sandlogic.com}}}

\date{\today}

\begin{document}
\maketitle



\begin{abstract}
We propose Samba ASR, the first state-of-the-art Automatic Speech Recognition (ASR) model leveraging the novel Mamba architecture as both encoder and decoder, built on the foundation of state-space models (SSMs). Unlike transformer-based ASR models, which rely on self-attention mechanisms to capture dependencies, Samba ASR effectively models both local and global temporal dependencies using efficient state-space dynamics, achieving remarkable performance gains. By addressing the limitations of transformers, such as quadratic scaling with input length and difficulty in handling long-range dependencies, Samba ASR achieves superior accuracy and efficiency. 

Experimental results demonstrate that Samba ASR surpasses existing open-source transformer-based ASR models across various standard benchmarks, establishing it as the new state-of-the-art in ASR. Extensive evaluations on the benchmark dataset show significant improvements in Word Error Rate (WER), with competitive performance even in low-resource scenarios. Furthermore, the inherent computational efficiency and parameter optimization of the Mamba architecture make Samba ASR a scalable and robust solution for diverse ASR tasks. 

Our contributions include the development of a new Samba ASR architecture for automatic speech recognition (ASR), demonstrating the superiority of structured state-space models (SSMs) over transformer-based models for speech sequence processing. We provide a comprehensive evaluation on public benchmarks, showcasing state-of-the-art (SOTA) performance, and present an in-depth analysis of computational efficiency, robustness to noise, and sequence generalization. This work highlights the viability of Mamba SSMs as a transformer-free alternative for efficient and accurate ASR. By leveraging the advancements of state-space modeling, Samba ASR redefines ASR performance standards and sets a new benchmark for future research in this field.

\end{abstract}


\section{Introduction}
The rapid evolution of deep learning has significantly transformed Automatic Speech Recognition (ASR), shifting from traditional systems such as Hidden Markov Models (HMMs) and Gaussian Mixture Models (GMMs) to advanced end-to-end neural architectures. While innovations such as Connectionist Temporal Classification (CTC) and attention-based encoder-decoder models have established new baselines \cite{graves2006connectionist} , transformer-based models like OpenAI’s Whisper have further pushed the boundaries, setting state-of-the-art benchmarks for multilingual, multitask ASR systems \cite{radford2022robustspeechrecognitionlargescale}. 

Despite their successes, transformer architectures face inherent challenges in scaling to long sequences, particularly those encountered in extended audio recordings. Transformers exhibit quadratic complexity with respect to sequence length, leading to high computational costs and memory usage for tasks requiring long-context modeling \cite{zuo2024falconmambacompetitiveattentionfree},\cite{glorioso2024zambacompact7bssm}. These limitations present a significant obstacle to achieving scalable and efficient ASR systems, especially in resource-constrained environments or for real-time applications. 

Structured State-Space Models (SSMs) \cite{gu2022efficientlymodelinglongsequences}  have emerged as a compelling alternative, offering efficient sequence modeling with linear complexity. The Mamba architecture \cite{gu2024mambalineartimesequencemodeling}, an innovation within this domain, extends SSM capabilities by introducing selective recurrence and hardware-aware optimizations. These advancements address the limitations of traditional linear time-invariant (LTI) dynamics, enabling Mamba to deliver exceptional efficiency and scalability. By leveraging selective state-space dynamics, Mamba achieves efficient long-range dependency modeling, making it particularly well-suited for ASR tasks. 

Mamba’s architecture introduces input-dependent parameters into the state-space equations, allowing for dynamic adaptation to sequence content. This capability compresses context into a smaller state representation while effectively capturing both local and global dependencies \cite{gu2024mambalineartimesequencemodeling}. Furthermore, Mamba employs hardware-aware techniques such as kernel fusion and parallel scan, optimizing computational efficiency and minimizing memory overhead during both training and inference. These features establish Mamba as a robust solution for sequence modeling across diverse modalities. 

While Mamba has demonstrated success in a range of applications, including language and vision tasks, its direct application to speech-to-text systems remained unexplored prior to this work. The development of Samba-ASR represents a significant breakthrough, showcasing the potential of Mamba-based architectures in ASR. By replacing traditional transformer encoders with Mamba’s efficient state-space modeling, Samba-ASR achieves state-of-the-art performance across major ASR benchmarks, including Gigaspeech \cite{chen2021gigaspeechevolvingmultidomainasr} and SPGISpeech \cite{oneill2021spgispeech5000hourstranscribed}. The model reduces inference latency and training time while maintaining high accuracy, even under challenging conditions such as noisy or spontaneous speech.  

The following sections delve into the technical details of State Space Models, the Mamba architecture, and its advancements in both language and vision tasks, setting the stage for our motivation and contributions to efficient ASR system using Mamba. 
    
\subsection{Background}
\subsubsection{State Space Models (SSMs)}
State Space Models (SSMs) \cite{gu2022efficientlymodelinglongsequences} provide a robust framework for sequence modeling by representing dynamical systems through a latent state that evolves over time. These models describe how inputs affect system states and how states generate output, using the following equations: 
\begin{center}
 \(h_{t+1} =A (h_t) +B (x_t)\) , \(y_t = C(h_t)\)   
\end{center}

where \(h_t\) is the latent state at time \(t\), \(x_t\)  is the input, \(y_t\) is the output, and \(A\), \(B\), \(C\) are parameter matrices. This formulation allows SSMs to efficiently model sequential data by transitioning between latent states and producing outputs influenced by both current and historical inputs. 

Traditionally, SSMs are linear time invariant (LTI), where \(A\), \(B\), \(C\) remain constant over time. Although LTI dynamics provides computational efficiency and stability, they limit the model's ability to adapt to input-dependent variations. Consequently, classical SSMs often struggle with complex, context-sensitive tasks, especially in discrete and content-rich modalities such as language. 

The matrices \(A\), \(B\),and \(C\) are learned parameters with the following interpretations.

\begin{itemize}
  \item \(A\): Determines how much the previous hidden state \(h_t\) should be considered to calculate the new hidden state \(h_{t+1}\).
  \item \(B\): Determines how much the input \(x_t\) should be considered to calculate the new hidden state \(h_{t+1}\).  
  \item \(C\): Determines how much the hidden state \(h_t\) should be considered in calculating the output \(y_t\).  
\end{itemize}

\subsubsection{Mamba: Linear-Time Sequence Modeling with Selective State Spaces}
Mamba\cite{gu2024mambalineartimesequencemodeling} extends traditional SSMs with a selectivity mechanism, addressing the limitations of LTI dynamics while preserving computational efficiency. Mamba’s formulation introduces input-dependent parameters into the state-space equations:	 

\begin{center}
 \(h_{t+1} =A (h_t) +B (x_t)\) , \(y_t = C(x_t)h_t\)   
\end{center}

where \(B (x_t)\) and \(C(x_t)\) are learned functions of the input \(x_t\) , allowing selective propagation of relevant information and enables dynamic adaptation to sequence content, while \(A\) remains a structured state transition matrix. This selective mechanism allows Mamba to efficiently compress context into a smaller state while maintaining the ability to capture long-range dependencies. 

To efficiently handle the introduced time-varying parameters, Mamba employs a hardware-aware implementation using techniques like kernel fusion, parallel scan, and recomputation. This minimizes memory overhead by leveraging GPU memory hierarchies, where state updates are computed in fast, low-level memory (e.g., SRAM) and final outputs are written to high-bandwidth memory (HBM). By avoiding the materialization of large latent states during training, Mamba achieves linear computational complexity while ensuring flexibility for diverse tasks. Furthermore, a recomputation strategy reduces memory requirements during backpropagation by recalculating intermediate states only when needed. 

The Mamba architecture simplifies traditional SSM designs by combining sequence transformation and gating mechanisms into a single homogenous block. This block replaces multi-head attention (MHA) and MLP components with a streamlined structure inspired by gated mechanisms in RNNs, such as:

  

\begin{center}
  \( g_t = \sigma(\text{Linear}(x_t)), \quad h_t = (1 - g_t) h_{t-1} + g_t x_t \)
\end{center}

where \(g_t\) represents the selection gate. By iteratively stacking these blocks with normalization (e.g., LayerNorm) and activation functions (e.g., SiLU), Mamba achieves high expressiveness while maintaining simplicity. Its design balances performance and efficiency, making it particularly effective for tasks such as Automatic Speech Recognition (ASR), language modeling, and reinforcement learning, where long-context dependencies and low latency are essential. 

\subsubsection{Advancements in Large Language and Vision Models Utilizing Mamba}
The Mamba architecture has inspired significant advancements in both language and vision modeling through its innovative state-space mechanism, leading to hybrid and pure Mamba-based models. 

Jamba\cite{lieber2024jambahybridtransformermambalanguage} introduces a novel hybrid architecture combining Transformer and Mamba layers, interleaved with mixture-of-experts (MoE) modules. This hybrid design addresses limitations of pure Transformer models in handling long contexts and computational efficiency. The resulting model, Jamba, achieves performance comparable to Mixtral-8x7B while supporting an unprecedented context length of 256,000 tokens—the longest among production-grade models. Jamba’s efficiency is remarkable, delivering three times the throughput of Mixtral-8x7B for long contexts and operating within a single 80GB GPU. This demonstrates the potential of integrating Transformer's attention mechanisms with Mamba’s efficient state-space dynamics for enhanced performance and resource utilization. 

Falcon Mamba\cite{zuo2024falconmambacompetitiveattentionfree} on the other hand, showcases the capabilities of a pure Mamba-based language model. This 7B parameter model trained on 5.8 trillion tokens challenges the notion that attention mechanisms are necessary for competitive performance. Surpassing open-weight Transformer-based models like Mistral 7B and Falcon2 11B, Falcon Mamba demonstrates that efficient inference and constant memory costs are achievable across context lengths. By addressing training stability issues with strategic initializations and RMSNorm placements, Falcon Mamba establishes itself as a competitive and efficient alternative to hybrid architectures. 

Zamba\cite{glorioso2024zambacompact7bssm} represents another leap in Mamba-based innovation by combining a Mamba backbone with a unique shared attention module. This 7B parameter model achieves competitive performance against leading transformer-based models while maintaining SSM efficiency. With faster inference speeds and reduced memory requirements, Zamba stands out as a resource-efficient model, particularly for generating long sequences. Although slightly behind in reasoning and in-context learning tasks due to limited training data, Zamba demonstrates the viability of hybrid SSM-attention designs for large-scale modeling. 

In vision tasks, Vision Mamba (Vim)\cite{zhu2024visionmambaefficientvisual} adapts Mamba for visual representation learning, demonstrating that self-attention mechanisms are not essential for effective vision modeling. Vim introduces bidirectional Mamba blocks to address positional awareness and global context challenges in vision tasks. The model delivers superior performance on benchmarks like ImageNet and COCO, achieving 2.8× faster inference speeds on high-resolution images compared to transformer-based models such as DeiT\cite{touvron2021trainingdataefficientimagetransformers}, while reducing GPU memory usage by 86.8\%. Vim’s sub quadratic computation and linear memory complexity make it a highly efficient solution for high-resolution visual tasks. 

These advancements illustrate the adaptability and efficacy of Mamba-based architectures in overcoming challenges across modalities, setting a new standard for resource-efficient and high-performing models in language and vision tasks. 

\subsection{Motivation}
Transformer-based ASR models, while successful, suffer from quadratic scaling, leading to high computational costs and memory usage when processing long audio sequences. This limitation becomes especially challenging with large datasets like Gigaspeech\cite{chen2021gigaspeechevolvingmultidomainasr} or SPGISpeech\cite{oneill2021spgispeech5000hourstranscribed}. To address these issues, we introduce Samba-ASR, which replaces the transformer encoder with the efficient Mamba SSM. The Mamba architecture offers linear complexity, allowing it to model long-range dependencies without the heavy computational burden. 

By leveraging Mamba’s selective state-space dynamics, Samba-ASR achieves state-of-the-art performance across major ASR benchmarks, surpassing transformer-based systems in both accuracy and efficiency. Our model reduces inference latency and training time, while maintaining robust performance even with noisy or spontaneous speech. Samba-ASR presents a scalable, efficient, and accurate solution for modern ASR tasks, setting a new standard in the field. 

\subsection{Contributions}
This paper makes the following key contributions: 

\begin{itemize}
  \item \textbf{Efficient ASR Architecture:}  We design Samba-ASR, integrating Mamba SSMs as encoders and decoder, achieving both accuracy and efficiency. 
  \item \textbf{SOTA Performance:} Samba-ASR achieves new benchmarks across Gigaspeech\cite{chen2021gigaspeechevolvingmultidomainasr}, LibriSpeech Clean/Other\cite{7178964LibriSpeech}, and SPGISpeech\cite{oneill2021spgispeech5000hourstranscribed}, outperforming existing transformer-based ASR systems.  
  \item \textbf{Efficiency Analysis:} Samba-ASR reduces both training time and inference latency, with linear scaling in sequence length 
  \item \textbf{Robustness}: Samba-ASR shows resilience to noisy and spontaneous speech, generalizing well across varied datasets.
\end{itemize}
Samba-ASR sets a new standard for efficiency and scalability in ASR systems, addressing critical challenges in modern speech recognition and paving the way for future innovations in the field.

\section{Related Work}
In recent years, Automatic Speech Recognition (ASR) systems have made significant strides in both accuracy and computational efficiency. Traditional models relied on recurrent and convolutional neural networks, but modern architectures, particularly those leveraging Transformer-based models, have set new benchmarks in performance. These Transformer models, such as Wave2Vec 2.0\cite{baevski2020wav2vec20frameworkselfsupervised}, Conformer\cite{gulati2020conformerconvolutionaugmentedtransformerspeech}, Whisper\cite{radford2022robustspeechrecognitionlargescale}, and Nvidia Canary\cite{puvvada2024moreaccuratespeechrecognition}, have greatly advanced ASR capabilities by capturing both local and global dependencies in speech data. However, despite their successes, these models often face challenges in terms of computational resources, scalability, and performance on long-form speech data. Recent innovations in State Space Models (SSMs), including the Mamba-based approaches, have emerged as promising alternatives, aiming to overcome these limitations. This section reviews the key developments in ASR technologies, discussing their strengths, limitations, and the contributions of the Mamba-based systems. 

\subsection{ Present ASR Systems}
\subsubsection{Wave2Vec 2.0}

The Wav2Vec2\cite{baevski2020wav2vec20frameworkselfsupervised} model is a widely adopted architecture for speech-to-text tasks, offering a robust method for processing raw audio into meaningful text. Its architecture comprises three main components: the feature encoder, quantization module, and Transformer encoder. The feature encoder processes raw audio waveforms using a series of convolutional layers that extract latent speech representations by down sampling the input while retaining critical temporal features. The quantization module discretizes these latent representations into a finite set of learned speech units using product quantization, which is crucial for self-supervised learning objectives. The Transformer encoder, a core part of the architecture, captures long-range dependencies in the audio data by contextualizing the extracted features through multi-layer attention mechanisms. During pretraining, a contrastive loss is employed by masking a portion of the feature encoder’s output and predicting the corresponding quantized representations, allowing the model to learn contextual speech representations effectively. In downstream tasks, such as speech-to-text generation, Wav2Vec2 is fine-tuned with labeled audio-text data, leveraging the Connectionist Temporal Classification (CTC) loss to map audio features directly to text sequences. This approach has demonstrated exceptional performance in automatic speech recognition (ASR), making Wav2Vec2 a foundational model in related works on ASR and audio-based sequence generation tasks. 

\subsubsection{Conformer}
The Conformer\cite{gulati2020conformerconvolutionaugmentedtransformerspeech} architecture has emerged as a significant advancement in speech processing models, particularly for Automatic Speech Recognition (ASR). It is designed to improve the extraction of both local and global features from audio signals by combining the strengths of convolutional networks and transformer-based attention mechanisms. This hybrid approach enables Conformer to achieve state-of-the-art performance in tasks requiring the understanding of sequential audio data, such as speech recognition. The core strength of the Conformer lies in its ability to effectively model both short-term and long-term dependencies, a challenge typically faced by traditional models relying on either convolutions or attention mechanisms alone. 

The preprocessing stage of the Conformer model begins with a convolutional subsampling layer. This initial step reduces the input sequence length by down sampling the feature maps, which not only reduces computational complexity but also retains essential information while discarding irrelevant details. The convolutional layer captures local patterns in the audio signal, which is crucial for preserving fine-grained temporal information. The output of this stage is then passed onto the main encoder, where the core feature extraction takes place. 

In the encoder, the audio data is processed by a sequence of Conformer blocks, each of which comprises four key modules: a feed-forward module (FFN), a multi-headed self-attention (MHSA) module, a convolution module, and a second FFN module. The MHSA module is responsible for capturing global contextual relationships within the input sequence, leveraging relative positional encoding to manage varying sequence lengths. This helps the model generalize better across different input sizes. The use of pre-norm residual connections in the MHSA module allows for stable and efficient training, as layer normalization is applied before the attention mechanism, followed by a residual connection that aids in gradient flow during training. 

The Conformer architecture combines convolutional and attention mechanisms to enhance speech recognition. By integrating these components, the model is able to handle varying input lengths while preserving both local and global features in the audio signal. The design, which uses a sandwich structure of different modules, helps balance feature extraction and computational efficiency. This makes Conformer a valuable approach for speech recognition tasks and other speech processing applications. 

\subsubsection{Whisper}
The Whisper model\cite{radford2022robustspeechrecognitionlargescale} is built on a sequence-to-sequence Transformer architecture, which is designed to handle various speech processing tasks such as transcription, translation, voice activity detection, and language identification. The input to the model is an 80-channel log-magnitude Mel spectrogram derived from raw audio, re-sampled at 16 kHz. The spectrogram is computed using 25-millisecond windows with a 10-millisecond stride, which captures the essential features of the audio signal. The model processes these features through a convolutional stem followed by a stack of Transformer blocks to learn meaningful representations of the speech signal. 

The encoder processes the Mel spectrograms through two initial convolutional layers followed by Transformer blocks. The convolution layers with GELU activation reduce the dimensionality of the spectrogram and capture local patterns, while the Transformer layers are responsible for extracting global temporal dependencies in the audio. The encoder also includes sinusoidal position embeddings, which help the model learn the temporal structure of the audio input. The encoder's output is a sequence of contextualized representations that capture the relevant acoustic and linguistic information from the audio. 

The decoder takes the encoder's output and generates text sequences, such as transcriptions or translations, depending on the task. It uses learned position embeddings and a set of special tokens to specify the task (e.g., transcription, translation). The decoder is trained to predict the next token in the sequence, conditioned on both the previously predicted tokens and the input audio features. The model is trained in a multitask setup, enabling it to perform multiple tasks like multilingual transcription and translation with a single unified architecture. The decoder ends with a special end of transcription token, marking the end of the output sequence. 

Thus, by using a Transformer-based architecture to handle various speech recognition tasks Whisper model processes Mel spectrograms through an encoder to capture audio features and then uses a decoder to generate text. This approach provides a unified solution for tasks like transcription and translation. 

\subsubsection{Nvidia Canary 1B}
The Canary model\cite{puvvada2024moreaccuratespeechrecognition} is an efficient encoder-decoder model designed for automatic speech recognition (ASR) and automatic speech translation (AST). It uses a FastConformer-based architecture, a speech-specific modification of the Conformer model, which balances high performance with reduced computational resources and training data. The model processes audio as Mel spectrograms, with a focus on minimizing the need for large datasets and achieves a 2.8x speedup over traditional models by increasing the down sampling factor to 8. 

The model employs a unified multi-task training strategy, where special prompt tokens direct it to perform either transcription or translation tasks. Canary is trained on synthetic data generated through machine translation, using advanced techniques such as dynamic data blending, data balancing, dynamic bucketing, and noise-robust fine-tuning. These methods optimize training efficiency, ensure consistent language representation, and minimize hallucinations when no speech is present. 

Despite being trained on just 86K hours of speech, much less than models like Whisper, which use up to 5M hours, Canary delivers competitive or superior performance. Its compact architecture and innovative training strategies make it highly effective for ASR and AST tasks, offering impressive results across multiple languages with significantly less training data. 

\subsection{Existing Mamba Based Approach }
Recent advancements in speech processing have been largely driven by Transformer-based\cite{vaswani2023attentionneed} models as discussed in the section 2.1, which excel at capturing global dependencies but face computational challenges for long-form sequences. State Space Models (SSMs), like Mamba, have emerged as efficient alternatives due to their linear computational scaling and ability to handle long-range dependencies. However, prior research, such as the BiMamba\cite{zhang2024mambaspeechalternativeselfattention} study, primarily focused on exploring bidirectional Mamba for tasks like speech enhancement and recognition without producing a standalone ASR system competitive with Transformer-based architectures. Similarly, "Exploring the Capability of Mamba in ASR"\cite{miyazaki2024exploringcapabilitymambaspeech} evaluated Mamba’s potential across various speech tasks, including ASR, text-to-speech, and summarization, showcasing comparable or superior performance to Transformer models like Conformer. However, this work remained domain-focused and did not result in a fully realized ASR model.

The "Speech Slytherin"\cite{jiang2024speechslytherinexaminingperformance} study extended Mamba's application to speech separation and synthesis, introducing hybrid models like Mamba-TasNet and ConMamba, which achieved competitive results but faced limitations in efficiency for shorter inputs and joint text-speech modeling. While these studies demonstrated Mamba’s promise in speech processing, none produced a robust ASR system capable of outperforming leading Transformer-based models. In contrast, our work introduces Samba-ASR, the first fully developed Mamba-based ASR system that surpasses Transformer architectures across major benchmarks, including Gigaspeech, LS Clean, LS Other, and SPGISpeech. This establishes Samba-ASR as a state-of-the-art solution, advancing the boundaries of speech recognition in terms of performance and computational efficiency. 

\section{Data processing}
The audio files are first loaded using the standard library torchaudio for efficient I/O operations. The audio file is decoded, down-mixed if necessary, and resampled to a fixed sample rate of 16 kHz, ensuring all audio inputs are in the same format, which is essential for uniform processing. Error handling is implemented to deal with any issues arising during the loading process, such as file format incompatibility or unsupported codecs. The loaded audio is then normalized to a range of [-1, 1] to facilitate model training. To ensure that the audio inputs match the expected size for processing, they are either padded or trimmed to a specific length \(N_{samples}\), defined by the model's requirements. This step is critical to maintain consistency in the length of audio segments processed by the encoder\cite{wang2024efficientwhisperstreamingspeech}. The choice of padding or trimming helps maintain the sequence length across all input samples, enabling efficient batch processing during training. Once the audio data is standardized, it is converted into a log-Mel spectrogram\cite{Koizumi_2018}, which captures frequency content and time dynamics. This is done by applying Short-Time Fourier Transform (STFT) \cite{kaneko2022istftnetfastlightweightmelspectrogram} to the audio waveform and projecting it onto the Mel filterbanks. The resulting magnitude spectrogram is then converted to a logarithmic scale to better match human auditory perception. This transformation enhances the discriminative power of the features, making them more suitable for speech recognition tasks. The spectrograms are further scaled to a range that ensures numerical stability and are normalized before being fed into the ASR model, facilitating accurate training and inference.

\subsection{Tokenizer}
The tokenizer is designed for the Mamba ASR (Automatic Speech Recognition) model, which converts textual input into a sequence of token IDs suitable for processing by the model. It includes a set of special tokens that mark the beginning and end of a transcription, indicate the task of transcribing the text, and potentially denote information for audio transcriptions. These tokens are crucial for guiding the model’s understanding of the input data. The tokenizer creates a basic vocabulary for English text that includes common ASCII characters, numbers, and punctuation marks. It uses a tokenizer of Byte level BPE (Byte Pair Encoding)\cite{sennrich2016neuralmachinetranslationrare} to segment the text into individual tokens. This method ensures that each element of the input text is represented uniformly, facilitating consistent preprocessing and accurate transcription when used with the Mamba ASR model.

\section{Samba-ASR: Architecture}
\subsection{Overview }
\begin{figure}[h!]
    \centering
    \includegraphics[width=14cm]{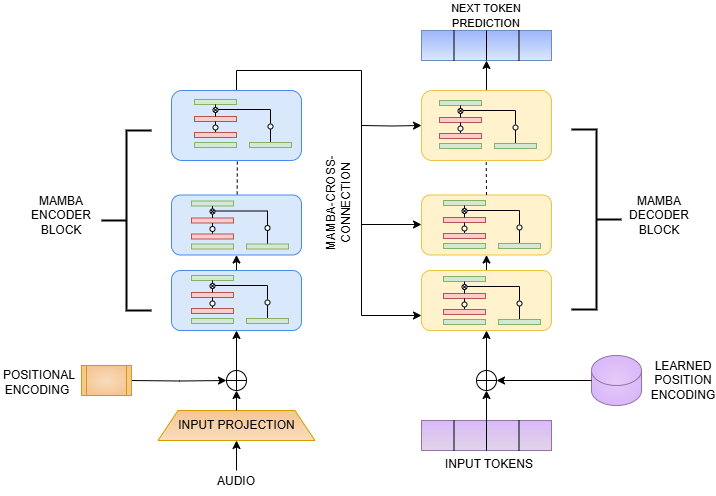}
    \caption{Architecture diagram (original) of the Samba-ASR model, illustrating the key components including the Mamba encoder, which processes raw audio features using Mamba blocks, and the Mamba decoder along with the Mamba-Cross-Connection bridge, which generates transcriptions by integrating audio context with text representations. The model's design focuses on efficient long-range dependency capture for accurate automatic speech recognition.}
    \label{fig:Model-architecture-diagram}
\end{figure}

Mamba ASR introduces a novel approach to Automatic Speech Recognition (ASR) by utilizing the Mamba architecture as shown in the figure \ref{fig:Model-architecture-diagram}, a state-of-the-art sequence modeling technique known for its computational efficiency and ability to capture long-range dependencies. Traditional Transformer-based models (e.g., Wav2Vec2 and Conformer) which predominantly use self-attention mechanisms for both audio feature extraction and text generation, Mamba ASR offers an alternative that uses state space models, allowing for better scalability and efficiency in processing longer sequences of data. This key distinction is central to Mamba ASR’s ability to handle both the audio and text components of ASR tasks more effectively. 

At the heart of Mamba ASR are two primary components: an audio encoder and a text decoder, both built with Mamba blocks. These blocks are designed to handle long-range dependencies in both speech and text sequences, offering a more efficient alternative to the memory-intensive approaches of Transformer and Conformer models. In contrast to these models, which use self-attention for global context capture (with varying computational efficiency), Mamba’s state space approach enables more efficient processing without sacrificing performance on tasks like transcription. 

\subsection{Encoder }
The Mamba ASR’s audio encoder processes raw audio input, represented as Mel spectrograms, to generate high-level feature representations that capture essential speech characteristics. It begins by passing the audio input through several convolutional layers\cite{gulati2020conformerconvolutionaugmentedtransformerspeech}, a technique borrowed from image processing models. These layers help to capture local temporal patterns in the audio signal, to preserve fine-grained details in audio features. The output of these convolutional layers is then passed through a series of Mamba blocks, which form the core of the encoder. 

Unlike Transformer-based models such as Wav2Vec2 and Conformer, which rely on self-attention mechanisms to capture global context, the Mamba encoder uses a state space model that scales linearly with sequence length, making it more computationally efficient for long audio sequences.  This results in a more efficient model for handling longer speech sequences without the quadratic complexity that Transformer-based models face. 

The output from the Mamba blocks is a sequence of contextualized audio embeddings, with Layer Normalization applied to stabilize the features before they are passed to the decoder. This efficient handling of long-range dependencies in audio sequences is critical for ASR tasks, where the model needs to capture and understand context across the entire utterance. 

\subsection{Decoder}
The text decoder in Mamba ASR generates the transcription from the encoded audio features. It begins by embedding the input tokens (representing the partially transcribed text) and adding positional embeddings to ensure the order of the sequence is preserved. These embeddings are then processed through a series of Mamba blocks, similar to the encoder. However, here, the decoder is conditioned on the encoded audio features via a Mamba-cross-connection mechanism. This allows the decoder to focus on the relevant portions of the audio sequence while predicting each token, which is essential for accurate transcription. 

In Transformer-based models like Wav2Vec2 and Whisper, the encoder directly feeds the decoder, and the self-attention mechanism captures the relationship between the audio features and the generated text. In contrast, Mamba ASR’s Mamba-cross-connection mechanism enables more targeted alignment between the audio and text features, improving the model’s ability to focus on specific audio segments that are most relevant to the current token being predicted. This targeted cross-connection mechanism helps the decoder refine the text representations, integrating both the audio context and previously predicted tokens. 

After passing through the Mamba blocks, a final Layer Normalization is applied, and the output is projected onto the vocabulary space via a linear layer followed by a softmax function\cite{ren2022fastspeech2fasthighquality}. This produces a probability distribution over the entire vocabulary, from which the model selects the most likely next token. To maintain the autoregressive nature of text generation, a causal mask ensures that predictions are based only on past tokens. 

The unique use of Mamba blocks in the decoder enables Mamba ASR to model the intricate relationship between audio features and text tokens effectively, addressing the complex alignment problem in ASR while also being computationally efficient. 



\section{Dataset}
To train Samba-ASR, we utilized a diverse set of high-quality speech datasets. The LibriSpeech clean split, containing 460 hours of transcribed 16kHz English speech, provided high-quality audio with minimal noise. We leveraged both the Train.100 (100 hours) and Train.360 (360 hours) subsets along with corresponding validation and test sets. These subsets include recordings with clear pronunciations and low Word Error Rates (WER), making them an ideal foundation for ASR training. 

Additionally, we incorporated the GigaSpeech dataset, which added 10,000 hours of transcribed audio from various sources such as audiobooks, podcasts, and YouTube. This dataset covers both read and spontaneous speaking styles across diverse topics including science and arts, enhancing the model’s ability to handle multi-domain speech and spontaneous variations in audio. 

We further enriched the training data with SPGISpeech, a domain-specific dataset consisting of 5,000 hours of transcribed financial audio. It features diverse accents (L1 and L2 English speakers), varying audio quality, and professionally formatted transcripts. This dataset played a crucial role in training Samba-ASR to excel in recognizing specialized financial terminologies and handling challenging audio conditions. 

\section{Training Details}

\begin{table}[!htbp]
 \setlength{\extrarowheight}{4pt}    
\centering
\begin{tabular}{|p{4cm}|p{4cm}|}
    \hline
    \multicolumn{2}{|c|}{\textbf{Training Parameters}} \\ \hline
    \textbf{Learning Rate}      & 1e-4    \\ \hline
    \textbf{Optimizer}      & AdamW   \\ \hline
    \textbf{Weight Decay }       & 0.01    \\ \hline
    \textbf{Adam eps }      & 1e-8    \\ \hline
    \textbf{Batch Size } & 256  \\ \hline

\end{tabular}

  \centering
  \vspace{4mm}
  \caption{Details of Training Parameters used for the training of Samba-ASR}
  \label{tab:training-parameters}
\end{table}

As detailed in Table \ref{tab:training-parameters}, the Samba-ASR model was trained with AdamW\cite{loshchilov2019decoupledweightdecayregularization} and gradient norm clipping along with a linear learning rate decay. A batch size of 256 was used, and the models were trained for 80 epochs with an initial learning rate of 1e-4, a weight decay of 0.01, and an Adam epsilon set to 1e-8. These parameters were selected to ensure stable convergence and effectively mitigate overfitting. Throughout the training process, we tracked training loss, validation loss, and Word Error Rate (WER) to monitor model performance and generalization. 

\begin{figure}[h!]
    \centering
    \includegraphics[width=10cm]{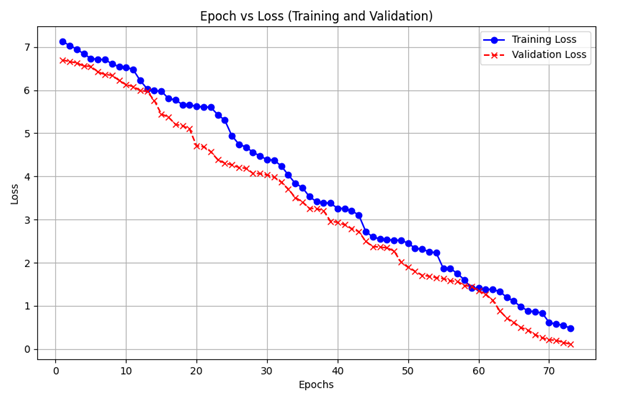}
    \caption{This graph shows the correlation of training and validation loss across epochs, with both losses steadily decreasing and converging around the 72nd epoch.}
    \label{fig:epoch_loss}
\end{figure}

\begin{figure}[h!]
    \centering
    \includegraphics[width=10cm]{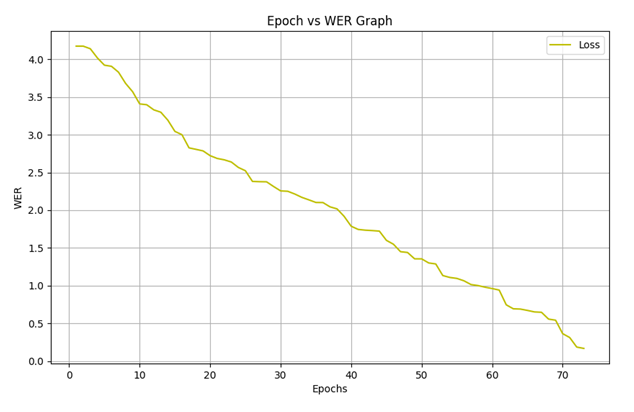}
    \caption{This graph demonstrates a significant reduction in Word Error Rate (WER) throughout the training process, indicating improved model performance and accuracy.}
    \label{fig:epoch_wer}
\end{figure}

As seen in the Epoch vs Loss graph as shown in the figure \ref{fig:epoch_loss}, both training and validation loss consistently decreased, starting from an initial value of approximately 7 and converging close to 0.5 by epoch 80. Similarly, the Epoch vs WER graph as shown in the figure  \ref{fig:epoch_wer} demonstrates a steady decline in WER, reducing from over 4.0 to approximately 0.2 by epoch 80. These results highlight the Samba-ASR model's ability to achieve stable convergence and significantly improve recognition accuracy, outperforming transformer-based ASR models. 

\section{Evaluation and Results}

We evaluate Samba-ASR (SandLogic) on four benchmark datasets—GigaSpeech, LibriSpeech (LS) Clean, LS Other, and SPGISpeech—and compare its performance with leading ASR models listed on the Open ASR Leaderboard hosted by Hugging Face. All results are computed using the same evaluation framework to ensure consistency and fairness. The primary evaluation metric is the Word Error Rate (WER). 

As shown in Table \ref{tab:Model-Performance-Comparison}, the model achieves a remarkable average WER of 3.65\%, outperforming top-performing systems. On LS Clean, it sets a new standard with a WER of 1.17\%, while maintaining a competitive edge on the more challenging LS Other subset with a WER of 2.48\%. Exceptional results are also observed on GigaSpeech and SPGISpeech, with WERs of 9.12\% and 1.84\%, respectively. These outcomes highlight the model's state-of-the-art performance and its ability to generalize effectively across diverse ASR benchmarks.

\begin{table}[!htbp]
\renewcommand{\arraystretch}{1.5} 
 \setlength{\extrarowheight}{4pt}    
\centering
\begin{tabular}{|l|c|c|c|c|c|}
    \hline
    \textbf{Model}                    & \textbf{Average WER} & \textbf{Gigaspeech} & \textbf{LS Clean} & \textbf{LS Other} & \textbf{SPGISpeech} \\ \hline
    Samba-ASR (SandLogic)     & \textbf{3.65}  & \textbf{9.12}                & \textbf{1.17} & \textbf{2.48}  & \textbf{1.84 }       \\ \hline
    nvidia/canary-1b  & \underline{4.15}   & 10.12 & 1.48 & 2.93 & \underline{2.06 }       \\ \hline
    nyrahealth/CrisperWhisper  & 4.69 & 10.24  & 1.82 & 4.00 & 2.7   \\ \hline
    nvidia/parakeet-tdt-1.1b & 7.01 & \underline{9.52} & 1\underline{.40} & \underline{2.60} & 3.16  \\ \hline
    openai/whisper-large-v3  & 7.44  & 10.02 & 2.01 & 3.91  & 2.94      \\ \hline
\end{tabular}

\centering
\vspace{4mm}
\caption{Model Performance Comparison Across Various Datasets}
\label{tab:Model-Performance-Comparison}
\end{table}

\section{ Conclusion}

Samba-ASR represents a significant breakthrough in automatic speech recognition technology, demonstrating superior performance across multiple benchmark datasets including GigaSpeech, LibriSpeech Clean/Other, and SPGISpeech. The model achieves remarkable results with an average Word Error Rate (WER) of 3.65\%, setting a new state-of-the-art benchmark with particularly impressive performance on LibriSpeech Clean (WER: 1.17\%) and SPGISpeech (WER: 1.84\%). 

The architecture's success can be attributed to its innovative use of state-space models (SSMs) in both encoder and decoder components, replacing traditional transformer-based attention mechanisms. This design choice results in linear computational complexity, enabling efficient processing of long audio sequences while maintaining high accuracy. The model's robust performance across diverse speaking styles, audio qualities, and domains demonstrates its practical viability for real-world applications. 

Samba-ASR's achievements extend beyond just performance metrics. The model's efficient architecture reduces both training time and inference latency, while maintaining linear scaling with sequence length. This combination of improved accuracy and computational efficiency establishes Samba-ASR as a compelling alternative to transformer-based models, setting a new direction for future research in speech recognition technology. 

\section{ Future Scope }
Future work on Samba-ASR will explore multiple key directions to enhance its capabilities, scalability, and broader applicability. A primary focus is extending support for multilingual ASR\cite{pratap2020massivelymultilingualasr50} and translation, enabling the system to process and transcribe speech in diverse languages, including those with limited resources. This will make Samba-ASR a robust tool for global applications, catering to cross-lingual communication and breaking language barriers effectively \cite{graves2013speechrecognitiondeeprecurrent}.  

To address diverse computational requirements, future iterations will explore the development of model variants with different sizes, from lightweight versions optimized for edge devices\cite{min2023macunifiedframeworkboosting} to larger, high-performance models for enterprise-level use. This scalability will ensure the system’s adaptability to various deployment scenarios, from real-time transcription on mobile devices to large-scale processing in cloud environments. 

Enhancing the encoder pre-training process is another critical avenue of research. By incorporating larger and more diverse datasets, we aim to further improve generalization across accents, dialects, and spontaneous speech variations. Additionally, integrating domain-adaptive fine-tuning will allow the model to excel in specific industries, such as healthcare or legal transcription. Finally, efforts to integrate real-time processing capabilities and on-the-fly language detection will make Samba-ASR even more versatile for dynamic and interactive use cases. These advancements will solidify Samba-ASR as a leading-edge solution in the ASR landscape, ensuring its continued evolution to meet emerging challenges in speech recognition.


\bibliographystyle{unsrt}  
\bibliography{main}

\end{document}